\documentclass[sigconf]{acmart}

\renewcommand\footnotetextcopyrightpermission[1]{} 
\usepackage{fancyhdr}
\usepackage[T1]{fontenc}
\usepackage{graphicx}
\usepackage{array}
\usepackage{float}
\usepackage{booktabs}
\usepackage{multirow}

\begin{document}

\title{An Embedding-based Approach to Inconsistency-tolerant Reasoning with Inconsistent Ontologies}

\author{Keyu Wang}
\affiliation{%
 \institution{ School of Computer Science and Engineering, Southeast University}
 \city{Nanjing}
 \country{China}}

\author{Site Li}
\affiliation{%
 \institution{School of Mathematics, Southeast University}
 \city{Nanjing}
 \country{China}}

\author{Jiaye Li}
\affiliation{%
 \institution{ School of Mathematics, Southeast University}
 \city{Nanjing}
 \country{China}}

\author{Guilin Qi}
\affiliation{%
 \institution{ School of Computer Science and Engineering, Southeast University}
 \city{Nanjing}
 \country{China}}

\author{Qiu Ji}
\affiliation{%
 \institution{School of Modern Posts \& Institute of Modern Posts, Nanjing University of Posts and Telecommunications}
 \city{Nanjing}
 \country{China}}

\begin{abstract}
Inconsistency handling is an important issue in knowledge management. Especially in ontology engineering, logical inconsistencies may occur during ontology construction. A natural way to reason with an inconsistent ontology is to utilize the maximal consistent subsets of the ontology. However, previous studies on selecting maximal consistent subsets have rarely considered the semantics of the axioms, which may result in irrational inference. In this paper, we present a novel approach to reasoning with inconsistent ontologies in description logics based on the embeddings of axioms. We first propose a sentence-embedding-based method and a knowledge-graph-embedding-based method for translating axioms into semantic vectors to calculate semantic similarities among axioms. We then define an embedding-based approach for selecting the maximal consistent subsets of the inconsistent ontology and use it to define an inconsistency-tolerant inference relation. We show the rationality of our inference relation by considering some logical properties. Finally, we conduct experiments on several ontologies to evaluate the effectiveness and efficiency of our proposed method. The experimental results show that our embedding-based method can outperform existing inconsistency-tolerant reasoning methods based on maximal consistent subsets.
\end{abstract}

\maketitle

\section{Introduction}
Ontologies are widely used in knowledge management and are critical for the success of the Semantic Web because they provide formal representation of knowledge shared within the Semantic Web applications. The development of the Semantic Web is further accelerated with the proposal of Knowledge Graph, which provides users with more intelligent services, such as more accurate recommendation and search \cite{1}. Ontologies also have a critical impact on the performance of Knowledge Graph reasoning \cite{2}. However, conflicting knowledge in ontologies is unavoidable. For example, ontology fusion \cite{3}, ontology evolution \cite{4} and ontology migration \cite{5} may result in inconsistent ontologies. Therefore, inconsistency handling is an essential issue in ontology engineering. 

A natural way to reason with inconsistent ontologies is to select maximal consistent subsets of an inconsistent ontology \cite{6}. An elementary method is called skeptical inference \cite{7}, i.e., an axiom can be inferred if it can be inferred from every maximal consistent subset of the inconsistent ontology. A well-known refinement of skeptical inference is to utilize the cardinality-maximal consistent subsets of the ontology for inference \cite{8}. However, in \cite{9}, the shortcomings of these two methods are pointed out. They fail to give fine consideration to the difference in the reliability of axioms, which results in weak reasoning power and poor answering quality. \cite{9} gives a general class of monotonic selection relations for comparing maximal consistent subsets. Each monotonic selection relation corresponds to a rational inference relation. However, the approaches given in \cite{9} are limited to propositional logic and it is not trivial to apply them to description logics. Axioms in an ontology contain semantic information, which can be used to define a rational inconsistency-tolerant reasoning method. We use an example to illustrate this. \\
\textbf{Example.} We consider an example of an inconsistent ontology that contains four axioms selected from  the widely used ontology OpenCyc\footnote{https://sourceforge.net/projects/opencyc/}:
\vspace{-0.4cm}
\begin{center}
    \begin{tabular}{l}
        $\varphi_1: ArtifactualFeatureType(Monument) $ \\
        $\varphi_2: ExistingStuffType(Monument) $  \\
        $\varphi_3: DisjointWith(ExistingObjectType, ExistingStuffType) $  \\
        $\varphi_4: SubClassOf(ArtifactualFeatureType, ExistingObjectType) $ 
    \end{tabular}
\end{center}

The existing methods based on propositional logic would assume that these axioms have the same status as shown in the example (continue) in Section 2, regardless of the semantics of axioms. However, these axioms contain different semantic information, which can be exploited to compute the reliability of axioms and select the maximal consistent subset of the ontology.

Pre-trained transformer-based language models for embedding such as BERT \cite{11} and Knowledge Graph Embedding models such as TransE \cite{transe} have been successfully applied in many natural language processing tasks and achieved good performance. Recently, they have been applied to ontology matching \cite{12} and instance matching \cite{13} to encode the semantics of instances or concepts in an ontology. They were also used to learn ontologies from knowledge graphs \cite{14}. This motivates us to use embedding-based models to encode the semantic information of an axiom in a description logics ontology and apply them to inconsistency-tolerant reasoning. 

In this paper, we propose a novel approach to reasoning with inconsistent OWL ontologies based on the embeddings of axioms. We first introduce two methods, a sentence-embedding-based method and a knowledge-graph-embedding-based method, for translating axioms into semantic vectors to compute semantic similarities among them. By considering the number of occurrences in the maximal consistent subsets and the degree of semantic relationships with other axioms, each axiom could be associated with a degree of reliability. We then propose an embedding-based approach to scoring the maximal consistent subsets and select the maximal consistent subset with the highest score. The selected maximal consistent subset of the ontology can be used to infer consistent and reasonable results. We show the rationality of our proposed reasoning method by considering some logical properties. Finally, we conduct experiments on several ontologies to evaluate the effectiveness and efficiency of our proposed method. The experimental results show that our embedding-based method can outperform existing inconsistency-tolerant reasoning methods based on maximal consistent subsets. The data, source codes and
technical report including proofs are provided in  the link: \href{https://github.com/sky-fish23/Embedding-based-infer}{https://github.com/sky-fish23/Embedding-based-infer}.

\vspace{-0.1cm}
\section{Related Work}
 There are mainly two classes of methods for inconsistency handling in DL-based ontologies. One deals with inconsistencies by repairing them and the other tolerates inconsistencies and changes the semantics of description logics. Our work falls into the latter class. In this part, we mainly discuss existing approaches to reasoning with inconsistent OWL DL ontologies based on consistent subsets. \cite{17,18} provide surveys on this topic. To find consistent subsets, \cite{19} proposes a linear extension strategy for checking whether an entailment could be inferred or not by defining syntactic relevance functions. This work is extended in \cite{20} by defining semantic relevance functions with Google distances. However, such an approach may not find maximal consistent subsets and it may result in many unknown answers to queries. 

To achieve some kind of minimal change in the calculation of the consistent subsets, researchers have proposed various methods to retain information as much as possible. The work in \cite{21} focuses on $\mathcal{SHIQ}$ ontologies, and assumes that an ontology is composed of a consistent TBox, and the instance assertions in the ABox are associated with weights. It answers a conjunctive query upon any subset of the set including the TBox and a weight-maximally consistent subset from the ABox. Similarly,  \cite{22} computes four kinds of maximal subsets by exploiting additional information in the ABox, which are maximal with respect to cardinality, weights, prioritized set inclusion or prioritized cardinality. They focus on DL-Lite, which is a sub-language of OWL DL. Furthermore, \cite{23} presents a sound and complete method for DL-Lite ontologies and an approximation method for more expressive DLs, so that the query answering systems can scale up to billions of data.  \cite{24} provides a practical approach by considering
three well-known semantics for DL-Lite ontologies and defines explanations for
answers. Recently,  \cite{25} proposes propositional encoding of maximality and then develops several SAT-based algorithms to calculate answers.

There are also some other methods to perform reasoning over an inconsistent ontology which do not rely on (maximal) consistent subsets of the ontology. For instance, various non-classical semantics could be adopted. The works in \cite{26,27} adopt four-valued and three-valued semantics for weakening an interpretation in DL from two truth values to four and three values respectively. To infer more useful information from an inconsistent ontology,  \cite{28} defines a novel description logics based on the quasi-classical logic. 

Different from all these existing methods, our approach does not rely on weight information to select maximal consistent subsets of an ontology as many real ontologies do not have weight information. Instead, we propose a new approach for selecting maximal consistent subsets of an ontology by considering axiom embedding.

\section{Preliminaries}
An ontology is a formal representation of knowledge that defines a set of concepts and relationships between them, and  OWL (Web Ontology Language) is a standard language recommended by W3C\footnote{https://www.w3.org/TR/owl-overview/} used to represent ontologies. Description logics (DLs), as the logic foundation of OWL, provide reasoning support for OWL ontologies and we refer to the DL Handbook \cite{15} for a detailed introduction. 

A DL-based ontology describes the characteristics of some properties and the relationships between entities through various axioms. Such axioms can be categorized into TBox (terminological axioms) and ABox (assertional axioms). A TBox establishes a conceptualization of a knowledge domain such as  concept inclusion in the form $C \sqsubseteq D$ , where $C$ and $D$ are concepts. An ABox describes particular individuals, such as concept assertions in the form of $C(a) $, where $C$ is a concept and $a$ is an individual.

An ontology is inconsistent if it has no model. To reason with an inconsistent ontology, we consider its maximal consistent sub-ontologies, which are maximal subsets of the ontology that are consistent. \\
\textbf{Definition 1 (MCS).} \textit{Given an ontology} $\mathrm{\Sigma}$, \textit{a maximal consistent sub-ontology(MCS)} $\mathrm{\Sigma^{\prime}}$ of $\mathrm{\Sigma}$ \textit{satisfies:} \\
\hspace*{0.2cm} $\bullet \ \
\mathrm{\Sigma^{\prime}} \subseteq \mathrm{\Sigma}$ \\
\hspace*{0.2cm} $\bullet \ \
\mathrm{\Sigma^{\prime}}$ \textit{is consistent} 	\\
\hspace*{0.2cm} $\bullet$ \ \ \textit{If} $ \mathrm{\Sigma^{\prime}} \subset \mathrm{\Sigma^{\prime \prime}} \subseteq \mathrm{\Sigma}$, \textit{then}  $\mathrm{\Sigma^{\prime \prime}}$ \textit{is not consistent}. 


We use mcs($\mathrm{\Sigma}$)  to represent the set composed of all the maximal consistent sub-ontologies of a given ontology $\mathrm{\Sigma}$.

\begin{figure*}
\centering
\includegraphics[width=0.87\textwidth]{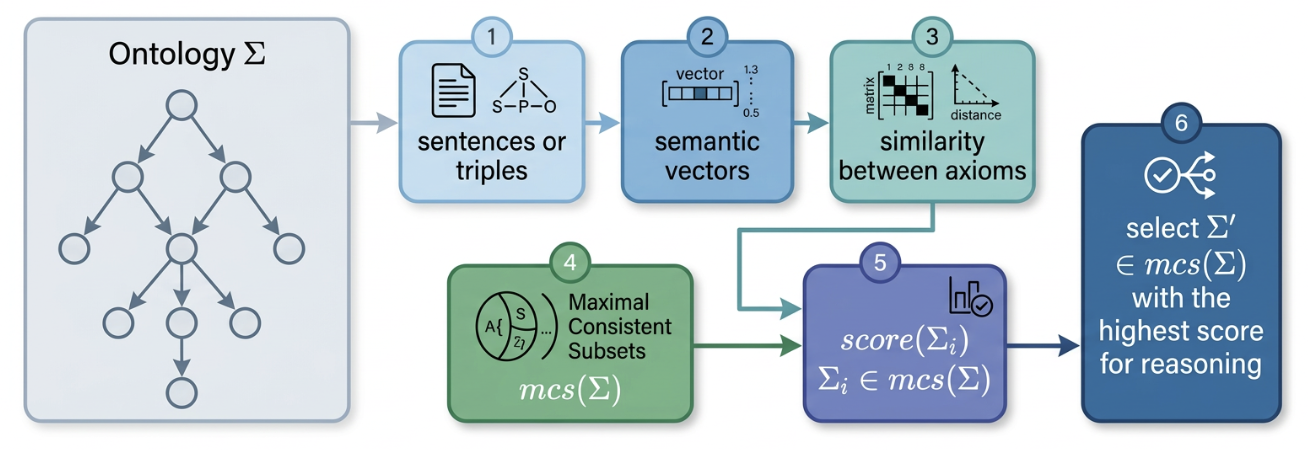}
\caption{\label{fig:method_figure} An embedding-based approach to reasoning with an inconsistent ontology.}
\end{figure*}

\cite{9} first defines mappings that attach a score to each axiom  of an ontology  and then aggregate those scores to rank each MCS. The MCS with the highest score is selected and used to define the inconsistency-tolerant reasoning method. Since only axioms are used when computing scores for axioms or MCSs,  we use the term ontology to refer to a (finite) set of axioms for simplification in the following definitions. In this paper, when we mention $\alpha \in \Sigma$ given an ontology $\Sigma$, we assume $\alpha$ is an axiom and  $|\mathrm{\Sigma}|$ denotes the number of axioms in the ontology $\mathrm{\Sigma}$. We adapt the notion of a scoring function given in \cite{9} to description logics.\\
\textbf{Definition 2} (scoring function). \textit{A} scoring function \textit{s associates with an ontology} $\mathrm{\Sigma}$ \textit{and an axiom} $ \alpha \in \mathrm{\Sigma} $ \textit{a non-negative real number} $s(\mathrm{\Sigma}, \alpha)$ which is equal to $0$ if and only if $\alpha$ is trivial (i.e., such that $\alpha$ is a tautology or a contradiction). 

 Different from \cite{9}, in DL ontologies in this paper, we exclude the cases where the axioms are trivial. \cite{9} designs several scoring functions, including the \#mc function mentioned in the following Definition 3. \\
\textbf{Definition 3} (\#mc). \textit{Let} $\mathrm{\Sigma}$ \textit{be an ontology and an axiom} $\alpha \in \mathrm{\Sigma}$. \textit{Define}:
$$
\setlength{\arraycolsep}{0.4pt}
\# mc(\mathrm{\Sigma},\alpha)= 
\left|\left\{\mathrm{\Sigma}_{i} \in mcs(\mathrm{\Sigma}) \mid \alpha \in \mathrm{\Sigma}_{i}\right\}\right|
$$ \\
\textbf{Definition 4} ($score^{\#mc}_{\mathrm{\Sigma},sum}$). \textit{Let}  $\mathrm{\Sigma}$  \textit{be an ontology and } $\mathrm{\Sigma}_i \in mcs(\mathrm{\Sigma})$. \textit{Define}: 
$$
\setlength{\arraycolsep}{0.4pt}
  score^{\#mc}_{\mathrm{\Sigma},sum}(\mathrm{\Sigma}_i) = \displaystyle\sum\limits_{\alpha \in \mathrm{\Sigma}_i} \#mc(\mathrm{\Sigma},\alpha) 
$$ 
\textbf{Example(continue)}  Ontology $\mathrm{\Sigma} = \{\varphi_1,\varphi_2,\varphi_3,\varphi_4 \}$.  $\mathrm{\Sigma}$ has 4 MCSs namely: 
$\{\varphi_2,\varphi_3,\varphi_4\}$, 
$\{\varphi_1,\varphi_3,\varphi_4\}$,
$\{\varphi_1,\varphi_2,\varphi_4\}$,
$\{\varphi_1,\varphi_2,\varphi_3\}$.
We have $\#mc(\mathrm{\Sigma},\varphi_1) = \#mc(\mathrm{\Sigma},\varphi_2) = \#mc(\mathrm{\Sigma},\varphi_3) = \#mc(\mathrm{\Sigma},\varphi_4) = 3 $, so the score of $\mathrm{\Sigma}_1(resp. \mathrm{\Sigma}_2, \mathrm{\Sigma}_3,\mathrm{\Sigma}_4)$ w.r.t $score_{\mathrm{\Sigma},sum}^{\#mc}$ is equal to 9 (resp. 9, 9, 9). These four MCSs all have the same value and are selected for reasoning, i.e., an axiom can be inferred if it can be inferred from each of these four MCSs.

Through the definitions above, the MCS $\mathrm{\Sigma}^{\prime}$ of the original inconsistent $\mathrm{\Sigma}$ with the highest score is selected, and the selected MCS can be used for reasoning. In the example above, all the four MCSs are selected because they all have the highest score, though this situation is rare in our proposed approach. In this work, we also define a reasoning method by attaching a score to each axiom in an ontology and scoring the MCSs. But different from \cite{9}, we propose an embedding-based novel method to score each axiom considering not only the syntax but also the semantics of the axioms.

In this work, a sentence-embedding-based method and a knowledge-graph-embedding-based method are proposed for translating axioms into semantic vectors to calculate semantic similarities among axioms. We introduce some background knowledge about embedding models as follows.\\
\textbf{Sentence Embedding.} A sentence embedding is a mathematical representation of a sentence in a fixed-length vector space that encodes the semantic meaning and contextual information of the sentence\cite{sesurvey}. Generally, the more similar the semantics of two sentences are, the closer their corresponding vectors are in the semantic space. In this work, we leverage Sentence-BERT\cite{30} and CoSENT\cite{consert} to get the semantic vectors of  axioms. Sentence-BERT trains the upper classification function by supervised learning and makes balances between performance and efficiency, while CoSENT takes advantage of a ranking loss function, which makes the training process closer to prediction.\\
\textbf{Knowledge Graph Embedding.}  Knowledge Graph Embedding (KGE) is a task for learning low-dimensional representation, typically called embeddings, of a knowledge graph’s entities and relations while preserving their semantics. A knowledge graph can be assumed as the ABox of an ontology while KGE can also be applied to embedding ontologies. For example, \cite{joie} uses KGE models such as TransE\cite{transe}, DisMult\cite{distmult} and HolE\cite{hole} to learn ontologies by jointly embedding instances and concepts. There are various KGE models and we refer to \cite{kge}  for a comprehensive survey. In this work, we use TransE \cite{transe} and RDF2Vec\cite{rdf2vec}. TransE is based on the assumption that $\mathbf{v_s} + \mathbf{v_r}$ is close to $\mathbf{v_o}$, where $\mathbf{v_s}, \mathbf{v_r}, \mathbf{v_o}$ are the embeddings of subject entity, relation and object entity respectively, while RDF2Vec utilizes random walks on the RDF graph to create sequences of RDF nodes, which are then used as input for the word2vec\cite{word2vec} algorithm.

\section{An Embedding-based Method to Reason with Inconsistent Ontologies}
The overall framework of our method is shown in Figure  \ref{fig:method_figure}. Our method can be divided into six parts. The first five parts can be assumed as offline preparation, including translating axioms into sentences or triples, using a sentence-embedding-based method or a knowledge-graph-embedding-based method to map the axioms into semantic vectors in a continuous space, calculating similarity between axioms based on the embeddings, generating and scoring the MCSs. Finally, we select the MCS with the highest score for reasoning.

\subsection{Semantic Representation \& Embedding}
We introduce two methods, a sentence-embedding-based method and a knowledge-graph-embedding-based method in this part, and either of the two methods can be used to convert axioms into semantic vectors. 

\begin{table*}[t]

\centering
\begin{tabular}{ p{4.0cm} cccc}
\toprule
&\multicolumn{2}{c}{}& \\
Ontology Elements &OWL Representation & Phrases or Sentences  & Triples \\
\midrule
\multirow{8}{*}{Concpets}&ObjectSomeValuesFrom(op A) & op at least one A & - \\
&ObjectAllValuesFrom(op A) & op only A & - \\ 
&ObjectHasValue(op a) & op a & - \\
&ObjectIntersectionOf(A B) & A and B & - \\
&ObjectUnionOf(A B) & A or B & - \\
&ObjectExactCardinality(n op A) & op exactly n A & - \\
&ObjectMinCardinality(n op A) & op at least n A & - \\
&ObjectMaxCardinality(n op A) & op at most n A & - \\
\midrule
\multirow{6}{*}{Axioms}
&SubClassOf(A B) & A is a kind of B & <A, SubClassOf, B> \\
&DisjointClasses(A B) & A isn’t a kind of B & <A, Disjointness, B>\\
&EquivalentClasses(A B) & A is a kind of B & <A, EquivalentClasses, B>\\
&ClassAssertion(A a) & a is a A & <a, isInstanceOf, C>\\
&ObjectPropertyAssertion(op a b) & a op b & <a, op, b>\\
&DataPropertyAssertion(dp a v) & a dp v & <a, dp, v>\\
\bottomrule
\end{tabular}
\caption{Rules to translate OWL concepts or axioms into sentences or triples, respectively, where \textit{A} and \textit{B} are concepts, \textit{a} and \textit{b} are individual names,  \textit{n} indicates an integer, \textit{op} and \textit{dp} indicate an object
property and a data property separately, and \textit{v} is a value.}
\label{table1}
\end{table*}

\textbf{Sentence-embedding-based method}. To represent the semantics of classes,  individuals and properties in an ontology, we first use NaturalOWL \cite{29} to translate axioms from OWL statements to natural language sentences. For example, the axiom ``ClassAssertion(ObjectMaxCardinality(1 madeFromGrape) 
 product145)'' is translated to ``product145 is a made from at most one Grape''. Table \ref{table1} shows most of the rules to translate OWL concepts or axioms into natural language sentences implemented by NaturalOWL.   Despite the fact that NaturalOWL does not translate all kinds of axioms or entities such as property inclusion axioms and transitive properties, the system can handle OWL statements in most ontologies, including various ontologies in our experiment.

Each axiom is transformed into a sentence in natural language form by NaturalOWL\cite{29}, and then the sentence is input into a certain sentence embedding model to obtain the corresponding semantic vector representation. Considering that BERT-based pre-trained language models have achieved good performance in text representation and text matching tasks, we use Sentence-BERT \cite{30} and CoSERT \cite{consert} as sentence embedding models to compute the embeddings of the axioms in natural language sentence form. We refer to the official website of PyPi\footnote{https://pypi.org/project/text2vec/} for more details about the implementation and experimental performance of these models.

\textbf{Knowledge-graph-embedding-based method}. The method based on sentence embedding can make use of the textual information of ontology, but ignores other semantic information of ontology, such as logical expression and class hierarchies. We propose another method using KG embedding, which is used to learn the relational information in the ontology. We implement the system ``TripleOWL'' of transforming axioms into triples and its transforming mechanism is similar to NaturalOWL. For example, an axiom ``SubClassOf(IceCream Food)'' is translated into ``<IceCream, SubClassOf, Food>''. The last column of Table \ref{table1} shows most of the rules to translate axioms to triples. Nonetheless, different from translating axioms to sentences, we fail to deal with complex concepts when translating axioms to triples, that is, we regard the complex concept in an axiom as a whole, which may cause worse performances for when the ontology has many axioms with complex concepts. We leave how to deal with complex concepts in KGE-based method to future work. The three parts of a triple are called subject entity, relation and object entity respectively for simplicity. Then the three components of the triples can be embedded into vectors using KG embedding models, and the vectors corresponding to the subject entity, relation and object entity are concatenated together to obtain the semantic vector representation of this axiom. We use TransE \cite{transe}, RDF2Vec \cite{rdf2vec} in our experiments, which are implemented by OpenKE platform \cite{openke} and pyRDF2Vec\cite{pyrdf2vec} respectively. 
\subsection{Semantic Similarity of Embedding}
We compute the semantic similarities among axioms. We denote the similarity of the axioms $\alpha$, $\beta$ as $Sim(\alpha,\beta)$ and the embedding of the axiom $\alpha$ as $Emb(\alpha)$. The similarity calculation method is given as follows:
\begin{center}
$ Sim(\alpha,\beta)  =Similarity(Emb(\alpha),Emb( \beta))$
\end{center} 

$Similarity$ represents certain similarity calculation method. The most common ones are based on Cosine Distance and Euclidean Distance:

\begin{center}  
$\textstyle Similarity_{Cos}(\mathbf{v_1}, \mathbf{v_2}) = \frac{1}{2} (1 + \frac{\mathbf{v_1} \cdot \mathbf{v_2}}{||\mathbf{v_1}|| \times ||\mathbf{v_2}||})$  \\   
$\textstyle Similarity_{Euc}(\mathbf{v_1}, \mathbf{v_2}) =  \frac{1}{1 + \sqrt{||\mathbf{v_1}-\mathbf{v_2}||}}$  \\  
\end{center}
where $\mathbf{v_1}$ and $\mathbf{v_2}$ are vectors of the same arity.

$Similarity_{Cos}$ and $Similarity_{Euc}$ satisfy the following three properties. For simplicity, we use $Sim(\phi , \varphi)$ to denote either $Similarity_{Cos}$  $(Emb(\phi) , Emb(\varphi))$ or $Similarity_{Euc}(Emb(\phi) , Emb(\varphi))$.\\
\textbf{Range.} The semantic similarity is a real number between 0 and 1: $0 \le Sim (\phi , \varphi) \le 1 $ for any $\phi$ and $\varphi$. The higher the similarity between $\phi$ and $\varphi$, the closer the $Sim (\phi , \varphi)$ is to 1, and the opposite is to 0. \\
\textbf{Grammatical Reflexivity.} Any axiom is always semantically closest to itself: $Sim (\phi, \phi) = 1$ for any $\phi$. \\
\textbf{Symmetry.} The semantic similarity between two axioms is symmetric: $Sim (\phi, \varphi) = Sim (\varphi, \phi)$ for any $\varphi$ and $\phi$. 

However, the similarity functions may not satisfy the semantic reflexivity defined below, because two semantically equivalent axioms may be converted into  different sentences. This does not affect that our proposed method satisfies the properties in Section 5 so our method still has the logical rationality. And in our work, we use a unique set of transformation rules proposed by NaturalOWL\cite{29}, and the experimental part shows the significant performance of our embedding-based method under this set of rules.\\
\textbf{Semantic Reflexivity.} Two semantically identical axioms should be closest to each other: If $ \models \phi \leftrightarrow  \varphi $, then $ Sim (\phi, \varphi)=1.$ \\
\textbf{Example (continue)} By applying the cosine similarity metric and Sentence-BERT, we obtain $Sim_{Cos}(\varphi_1, \varphi_2) = 0.41$ and $ Sim_{Cos}(\varphi_1, \varphi_4)$ $ = 0.62 $. This may be because $\varphi_4$ has a same concept \textit{ArtifactualFeatureType} with $\varphi_1$ while $\varphi_3$ doesn't, so $\varphi_4$ has higher semantic association with $\varphi_1$.

\subsection{Semantic Selection Functions}
We first define the degree of aggregation of each axiom in the MCS. Then we  score each axiom to represent the reliability of the axioms. Finally we aggregate the scores of the axioms to get the score of each MCS.

We use the degree of aggregation to express how closely an axiom relates to other axioms in a MCS. The greater the aggregation degree of an axiom in a MCS, the closer the axiom is to the semantics of other axioms in this MCS, which indicates more semantic information it contains about this MCS.  \\ 
\textbf{Definition 5}  (agg). \textit{Given an ontology} $\mathrm{\Sigma}$ \textit{and} $\mathrm{\Sigma}_i \in mcs(\mathrm{\Sigma})$, \textit{we define the aggregation of an axiom} $\alpha \in \mathrm{\Sigma}_i $ \textit{as follows}: 
\begin{center}
    $ agg(\mathrm{\Sigma}_i, \alpha) = \displaystyle\frac{1}{|\mathrm{\Sigma}_i|} \displaystyle\sum\limits_{\beta \in \mathrm{\Sigma}_i} Sim(\alpha, \beta) $
\end{center} 

We then calculate the score for each axiom. If an axiom exists in more MCSs and it has a higher degree of aggregation in the MCSs it appears in which means it is more likely a true axiom in these MCSs,  this axiom is considered to be more reliable. Below, we define scoring function for axioms.\\
\textbf{Definition 6} (mc). \textit{Given an ontology} $\mathrm{\Sigma}$ \textit{and an axiom} $\alpha \in \mathrm{\Sigma}$, \textit{we define the score of an axiom} \  $\alpha \in \mathrm{\Sigma}$ \ \textit{as follows}:
\begin{center}
    $mc(\mathrm{\Sigma}, \alpha) = \ \displaystyle\sum\limits_{\{\mathrm{\Sigma}_i \in mcs(\mathrm{\Sigma}) | \alpha \in \mathrm{\Sigma}_i \}} agg(\mathrm{\Sigma}_i, \alpha)$
\end{center}

Finally we accumulate the scores of axioms in each MCS to obtain the scores of the MCSs.\\
\textbf{Definition 7} (scoring function) \textit{Given an ontology} $\mathrm{\Sigma}$ \textit{and} $\mathrm{\Sigma}_i \in mcs( \mathrm{\Sigma} )$, \textit{the scoring function for the maximal consistent sub-ontologies are defined as follows}:
\begin{center} 
     $score(\mathrm{\Sigma}_i) = \displaystyle\sum\limits_{\alpha \in \mathrm{\Sigma}_i} mc(\mathrm{\Sigma}, \alpha)$
\end{center}
 
Each MCS is assigned a score by the scoring function in Definition 7. Then we select the MCS with the highest score for reasoning.   \\
\textbf{Example (continue).} For the  example mentioned above, we summarize the above calculations: MCS $\Sigma_3 = \{\varphi_1, \varphi_2, \varphi_4\}$ gets the highest score.  In our method, we conclude $\varphi_1, \varphi_2, \varphi_4$ and use these axioms for reasoning. \\

\section{Logical Properties}
We consider the logical properties of the inference relation between the axioms in $\Sigma$ and reasoning results using our method. An inference relation is rational when it satisfies the six properties in the minimal set of expected properties of preferential inference relations  (also called \textit{system} P) and  one of the rational inference relations  (also called \textit{system} R) \cite{7}.   \\
$ \hspace*{1.2cm}   \textbf{Ref} \quad \displaystyle\alpha \mid\sim \alpha \quad  \quad \quad  \quad \quad \ \  \  \
\textbf{Cut} \quad  \displaystyle\frac{\alpha \wedge \beta \mid\sim \gamma, \alpha\mid\sim \beta}{\alpha \mid \sim \gamma}  \\
  \hspace*{1.2cm} \textbf{LLE} \quad \displaystyle\frac{\vDash \alpha \leftrightarrow \beta, \alpha \mid\sim \gamma}{\beta \mid \sim \gamma} \quad   \ \ \
  \textbf{Or} \quad \displaystyle\frac{\alpha|\sim \gamma, \beta\mid\sim \gamma}{\alpha \vee \beta \mid \sim \gamma}\\ 
   \hspace*{1.2cm} \textbf{RW} \quad \displaystyle\frac{\vDash \alpha \rightarrow \beta, \gamma \mid \sim \alpha}{\gamma \mid \sim \beta} \quad \ \ \
  \textbf{CM} \quad \displaystyle\frac{\alpha\mid\sim \beta, \alpha\mid\sim \gamma}{\alpha \wedge \beta \mid \sim \gamma}  \\
  \hspace*{1.2cm} \textbf{RM} \quad \displaystyle\frac{\alpha\mid \not\sim \neg \beta, \alpha \mid\sim \gamma}{\alpha \wedge \beta \mid \sim \gamma}$

For example, \textbf{Cut} expresses the fact that one may, in his way towards a plausible conclusion, first add a hypothesis to the facts he knows to be true and prove the plausibility of his conclusion from this enlarged set of facts, and then deduce (plausibly) this added hypothesis from the facts. We refer to the technical report\footnote{https://github.com/sky-fish23/Embedding-based-infer/blob/main/technical\%20report.pdf} for detailed explanations about these properties.

To investigate whether our inference relation satisfies the logical properties above, we present some significant definitions in \cite{7} and theorems as follows: \\
\textbf{Definition 8} (Aggregation function). $\oplus$ \textit{is an aggregation function if for every positive integer n, for every non-negative real number} $x_1, ... x_n$, $\oplus(x_1, ... x_n)$ \ \textit{is a non-negative real number}. 

For example, this paper chooses \textit{sum} as the aggregation function.\\
\textbf{Definition 9} ($score^{s}_{\Sigma,\oplus}$). \textit{Let} $\mathrm{s}$ \textit{be a scoring function  defined in Definition 2 and let } $\oplus$ \textit{be an aggregation function}. \textit{Let} $\mathrm{\Sigma}$ \textit{be an ontology and} \  $\mathrm{\Sigma}_i = \{\alpha_1 ... \alpha_n\} \subseteq \mathrm{\Sigma} \ $. \ \textit{We define} $score^{s}_{\mathrm{\Sigma}, \oplus} (\mathrm{\Sigma}_i) = \oplus_{\alpha \in \mathrm{\Sigma}_i} s(\mathrm{\Sigma}, \alpha)$.

In Definition 7, we denote $score^{mc}_{\mathrm{\Sigma}, sum} (\mathrm{\Sigma}_i)$ as $score(\mathrm{\Sigma}_i)$ for simplicity. On the foundation of Definition 9, we can define the monotonic selection relation according to our method. Before that, we propose a more general one. \\
\textbf{Definition 10} (Monotonic selection relation). \textit{Given an ontology}  $\mathrm{\Sigma}$, \textit{let}  $\succeq_{\Sigma} \ \subseteq 2^{\Sigma} \times 2^{\Sigma}$ \textit{ be a reflexive, transitive and total relation over the powerset of}  $\mathrm{\Sigma}$. $\succeq_{\Sigma}$  \textit{is said to be a monotonic selection relation if for every  consistent set} \ $\mathrm{\Sigma}_i \subseteq \mathrm{\Sigma}$ , \textit{for every non-trivial axiom}  $\alpha \in \mathrm{\Sigma} \backslash \mathrm{\Sigma}_{i}, \mathrm{\Sigma}_{i} \cup\{\alpha\} \succ_{\Sigma} \mathrm{\Sigma}_{i}$. 

For instance, the relation $\succeq_{card}$ defined over $P(\Sigma)$ is a monotonic selection relation, in which $\Sigma_i \succeq_{card} \Sigma_j$ if and only if $|\mathrm{\Sigma}_i| \ge |\mathrm{\Sigma}_j|$. To be specific, a selection relation and an inference relation in \cite{9} are given as follows:\\
\textbf{Definition 11} $\left(\succeq_{\Sigma, \oplus}^{s}\right)$. \textit{Let s be a scoring function and} $\oplus$ \textit{an aggregation function. Let} $\mathrm{\Sigma}$ \textit{be an ontology}, $\mathrm{\Sigma}_{i}, \mathrm{\Sigma}_{j} \subseteq \mathrm{\Sigma}$. \textit{We state that} $\mathrm{\Sigma}_{i} \succeq_{\mathrm{\Sigma}, \oplus}^{s} \mathrm{~\Sigma}_{j}$ \ \textit{if and only if} $\mathrm{score}_{\mathrm{\Sigma}, \oplus}^{s}\left(\mathrm{~\Sigma}_{i}\right) \geq$ $\operatorname{score}_{\mathrm{\Sigma}, \oplus}^{s}\left(\mathrm{~\Sigma}_{j}\right)$.

Based on Definition 11, we can compare the subsets of an ontology by the scores of them. We adopt the following notation for convenience: $\operatorname{mcs}(\mathrm{\Sigma}, \alpha)=\{\mathrm{\Sigma}_i \subseteq \mathrm{\Sigma} \mid \mathrm{\Sigma}_i \cup\{\alpha\}\in \mathrm{mcs}(\mathrm{\Sigma} \cup\{\alpha\})\} $.\\
\textbf{Definition 12} ($mcs_{\succeq_\mathrm{\Sigma}}$). \textit{Given an ontology} \ $\mathrm{\Sigma}$, \ \textit{an  axiom} \ $\alpha \in \mathrm{\Sigma}$, \textit{and a monotonic selection relation} \ $\succeq_\mathrm{\Sigma}$, \textit{we define} $ mcs_{\succeq_\mathrm{\Sigma}}(\mathrm{\Sigma}, \alpha) = \{ \mathrm{\Sigma}_i \in mcs(\mathrm{\Sigma}, \alpha) |  \textit{there exists}\textit{no} \  \mathrm{\Sigma}^{\prime}_i \in mcs(\mathrm{\Sigma}, \alpha) \ \textit{such that} \ \mathrm{\Sigma}^{\prime}_i \succ_{\mathrm{\Sigma}} \mathrm{\Sigma}_i \}$. 

With respect to a monotonic selection relation, we define a selection mechanism consisting in keeping only the best subsets.\\
\textbf{Definition 13} \cite{9} $\left(\mid \sim_{\Sigma}^{{mcs}_{\succeq \Sigma}}\right.$, \textit{a binary inference relation based on best subsets  w.r.t} $\left.\succeq_{\Sigma}\right)$. \textit{Given an ontology} $\mathrm{\Sigma}$, \textit{two axioms} $\alpha$ \textit{and} $\beta$, \textit{and a monotonic selection relation} ${\succeq_\Sigma}$, \textit{we state that} $\alpha \mid \sim_{\Sigma}^{{mcs}_{\succeq \Sigma}} \beta$ \textit{if and only if }  \textit{for every} $\Sigma_{i} \in mcs_{\succeq_{\Sigma}}(\Sigma, \alpha)$, \textit{we have} $\Sigma_{i} \cup\{\alpha\}  \models \beta$. 

We propose the inference relation to show how conclusions are inferred from an ontology. \\
\textbf{Theorem 1}. Let $\Sigma$ be an ontology and let $\succeq_{\Sigma}$  be a selection relation over $\Sigma$. Then $\succeq_{\Sigma}$  is monotonic if and only if there exists a scoring function \textit{s} for $\Sigma$ such that $s(\Sigma, \alpha) > 0 $ for every axiom $\alpha \in \Sigma$. \\
\textbf{Proof sketch of Theorem 1}. From the definition of monotonic selection relation and $\succeq_{\oplus, \Sigma}$, this result can be easily induced. Especially, for necessity, to ensure the arbitrariness of $\alpha$, we have to show that for every axiom $\alpha \in \Sigma$, there exists a consistent subset $\Sigma_i \subseteq \Sigma$, such that $\alpha \notin \Sigma_i $.

Based on Theorem 1, we show that the given selection relation is monotonic. Furthermore, Theorem 2 shows the equivalence of the rationality of an inference relation and the monotonicity of its corresponding selection relation. \\
\textbf{Theorem 2}. \textit{A relation} $ \mid\sim $ \textit{is rational if and only if there exists an ontology $\mathrm{\Sigma}$ and a monotonic selection relation} $ \succeq_{\Sigma} $ \textit{such that} $ \mid \sim_{\Sigma}^{{mcs}_{\succeq \Sigma}} \ = \ \mid\sim $. \\
\textbf{Proof sketch of Theorem 2}. Theorem 2 is a rewriting of Theorem 5.18 in \cite{7}, so the proof is the same as that.\\
\textbf{Theorem 3}. \textit{The relation} $ \mid\sim $ \textit{defined by the selection relation which takes our proposed scoring function is rational}. \\
\textbf{Proof sketch of Theorem 3.} According to Theorem 2, we know that the rationality of the inference relation is equivalent to the monotonicity of corresponding selection relation. So we have to verify the positivity of our scoring functions to ensure every axiom $\alpha \in \mathrm{\Sigma} , s(\mathrm{\Sigma}, \alpha) > 0$ (We exclude the trivial axioms).

In conclusion, the selection relation based on our proposed scoring function is monotonic relation according to Theorem 1. And Theorem 2 shows the rationality of the corresponding inference relation. Due to the rationality of our proposed method, the reasoning satisfies all the seven logical properties mentioned above. Detailed proofs of these theorems and explanation of the properties can be found in the technical report
\footnote{\href{https://github.com/sky-fish23/Embedding-based-infer/blob/main/technical\%20report.pdf}{https://github.com/sky-fish23/Embedding-based-infer/blob/main/technical\%20report.pdf}}.

\section{Experiment and Evaluation}
In this section, we first introduce the experimental dataset and settings. Then we conduct some experiments to show the performance of our proposed method.

\begin{table}[t]
    \centering
    \setlength{\tabcolsep}{0.2mm}{
    \begin{tabular}{c|cccccc}
    \hline
    Ontology Name & \textbf{\#class} & \textbf{\#prop.} & \textbf{\#indi.} & \textbf{\#axiom} & \textbf{\#MCS} & \textbf{Expressivity}  \\ \hline
    AUTO.-cocus-eda & $158$ & $86$ & $114$ & $907$ & $6$ & $\mathcal{ALCOIN}$  \\
    bioportal-metadata & $98$ & $183$ & $61$ & $822$ & $14$ & $\mathcal{ALUHIN}$ \\
    UOBM-lite-10-35 & $52$ & $40$ & $43$ & $162$ & $16$ & $\mathcal{SHOIN}$ \\
    UOBM-lite-10-36 & $52$ & $40$ & $43$ & $163$ & $40$ & $\mathcal{SHOIN}$ \\
    \hline
    \end{tabular}}
    \caption{ Inconsistent ontologies used in the evaluation.}
    \label{tab:my_label}
\end{table}

\subsection{Data}
In order to fairly evaluate the quality of our proposed method in answering queries, the test ontology dataset should be diverse. We select four inconsistent ontologies, which vary in sources, size, expressiveness, number of axioms and MCSs. Table \ref{tab:my_label} provides details of the dataset, where \#class, \#prop., \#indi., \#axiom and \#MCS represent the number of classes, properties, individuals, axioms and MCSs for the ontology respectively. Half of the ontologies are existing inconsistent ontologies crafted by others, and others are created by us based on existing ontologies.

AUT.-cocus-eda is constructed in \cite{kbs14} by merging two source ontologies \textit{cocus} and \textit{edas} with their mapping generated by the mapping system AUTOMSv2 which has participated in the famous contest of ontology alignment evaluation initiative\footnote{ http://oaei.ontologymatching.org/2012/conference/}. 
Its inconsistency is caused by multiple sources, lying on two similar concepts `cocus\#person' and `edas\#person'. Bioportal-metadata\footnote{ https://bioportal.bioontology.org/} is a real-life ontology from the world’s most comprehensive repository of biomedical ontologies \cite{kbs14}. Some inconsistencies come from multiple sources of certain individual. For instance, these four axioms are inconsistent: \\
\begin{small}
$ \varphi_1:  SubClassOf(KnowledgeRepresentationParadigm,$ \\ 
\hspace*{2.4cm} $ documentation \ max \ 1 \ rdfs:Literal)$ \\
${\varphi_2: Domain(documention, KnowledgeRepresentationParadigm)}$ \\
${\varphi_3: documentation(Jena\text{-}ARQ, ``http://jena.sourceforge.net/ARQ/")}$ \\
${\varphi_4: documentation(Jena\text{-}ARQ,``http://jena.sourceforge.net/")}$ 
\end{small}

The two UOBM ontologies are constructed by us based on UOBM-lite-10 with an increasing number of axioms. UOBM \cite{31} indicates University Benchmark and is enriched from the famous Lehigh University Benchmark (LUMB) \cite{32}. UOBM-lite-10 means that it is enriched from LUBM by adding OWL Lite constructors and contains individuals from 10 universities. To construct inconsistent ontologies, we borrowed the idea from \cite{21} and develop the tool \textit{Injector}\footnote{https://anonymous.4open.science/r/Embedding-based-infer/data/Injector/} to insert different numbers of conflicts. The mechanism to generate conflicts in this work is the same as \cite{21} and details about the implementation of \textit{Injector} are shown in section 5 of \cite{21}. A conflict is a set of instance assertions violating a functional role restriction or a disjointness constraint. UOBM-lite-10-35 and UOBM-lite-10-36 are obtained by inserting 35 and 36 conflicts into ontology UOBM-lite-10 respectively. 

\begin{table*}
    \begin{center}
     \resizebox{0.8\hsize}{!}{
     \begin{tabular}{|c|c|c|c|c|c|c|c|c|c|}
    \hline
        \textbf{ Method } & \textbf{ Similarity  } &
        \textbf{ Queries } & \textbf{ IA  } & \textbf{  CA } &
        \textbf{RA} &
        \textbf{CIA} &\textbf{IA Rate} &\textbf{ICR Rate}  \\
        \hline
        \multicolumn{9}{|c|}{AUT.-cocus-edas}   \\ \hline
       \multicolumn{2}{|c|}{Skeptical inference}& 124 & 6 & 118 & 0 & 0 & 4.84\%  & 100\%\\ \hline
       \multicolumn{2}{|c|}{CMCS} & 124 & 8 & 116 & 0 & 0 & 6.45\% & 100\%\\ \hline
       \multicolumn{2}{|c|}{\#mc} & 124 & 8 & 116 & 0 & 0 & 6.45\% & 100\%\\ \hline
       {Sentence-BERT} &  
       Cosine  & 124 & 122 & 0 & 2 & 0 & 98.39\% & 100\% \\ \cline{2-9}
        & Euclid  & 124 & 122 & 0 & 2 & 0 & 98.39\% & 100\% \\
        \hline
       {ConSERT} & Cosine & 124 & 119 & 3 & 1 & 1 & 95.97\% & 99.21\% \\ \cline{2-9}
        & Euclid & 124 & 122 & 0 & 3 & 0 & 98.39\% & 100\% \\
        \hline
       {TransE} & Cosine   & 124 & 8 & 116 & 0 & 0 & 6.45\% & 100\%\\ \cline{2-9}
        & Euclid  & 124 & 8 & 116 & 0 & 0 & 6.45\% & 100\%\\ 
        \hline
       {RDF2Vec} & Cosine & 124 & 122 & 0 & 2 & 0 & 98.39\% & 100\%\\ \cline{2-9}
        & Euclid & 124 & 8 & 116 & 0 & 0 & 6.45\% & 100\%\\ \hline
        \multicolumn{9}{|c|}{Bioportal metadata}   \\ \hline
       \multicolumn{2}{|c|}{Skeptical inference} & 41 & 22 & 19 & 0 & 0 & 53.66\%  & 100\% \\ \hline
       \multicolumn{2}{|c|}{CMCS} & 41 & 22 & 19 & 0 & 0 & 53.66\%  & 100\% \\ \hline
       \multicolumn{2}{|c|}{\#mc} & 41 & 22 & 19 & 0 & 0 & 53.66\%  & 100\% \\ \hline 
       {Sentence-BERT} &  
       Cosine & 41 & 36 & 0 & 3 & 2 & 87.80\% & 95.12\% \\ \cline{2-9}
        & Euclid & 41 & 38 & 0 & 3 & 0 & 92.68\% & 100\% \\
        \hline
       {ConSERT} & Cosine & 41 & 38 & 0 & 3 & 0 & 92.68\% & 100\% \\ \cline{2-9}
        & Euclid & 41 & 38 & 0 & 3 & 0 & 92.68\% & 100\% \\
        \hline
       {TransE} & Cosine  & 41 & 38 & 0 & 3 & 0 & 92.68\% & 100\% \\ \cline{2-9}
        &Euclid & 41 & 38 & 0 & 3 & 0 & 92.68\% & 100\% \\
        \hline
       {RDF2Vec} & Cosine & 41 & 36 & 2 & 2 & 1 & 87.80\% & 97.56\% \\ \cline{2-9}
        & Euclid & 41 & 22 & 19 & 0 & 0 & 53.66\% & 100\% \\ \hline
        \multicolumn{9}{|c|}{UOBM-lite-10-35}   \\ \hline
       \multicolumn{2}{|c|}{Skeptical inference} & 90 & 74 & 16 & 0 & 0 & 82.22\%  & 100\% \\ \hline
       \multicolumn{2}{|c|}{CMCS} & 90 & 74 & 16 & 0 & 0 & 82.22\%  & 100\% \\ \hline
       \multicolumn{2}{|c|}{\#mc} & 90 & 74 & 16 & 0 & 0 & 82.22\%  & 100\% \\ \hline 
       {Sentence-BERT} &  
       Cosine & 90 & 80 & 0 & 4 & 6 & 88.89\% & 93.33\% \\ \cline{2-9}
        & Euclid & 90 & 89 & 0 & 0 & 1 & 98.89\% & 98.89\% \\
        \hline
       {ConSERT} & Cosine & 90 & 84 & 0 & 4 & 2 & 93.33\% & 97.78\% \\ \cline{2-9}
        & Euclid & 90 & 89 & 0 & 0 & 1 & 98.89\% & 98.89\% \\
        \hline
       {TransE} & Cosine & 90  & 83 & 1 & 4 & 2 & 92.22\% & 97.78\% \\ \cline{2-9}
        & Euclid & 90 & 89 & 0 & 0 & 1 & 98.89\% & 98.89\% \\
        \hline
       {RDF2Vec} & Cosine & 90 & 78 & 3 & 4 & 5 & 87.67\% & 94.44\% \\ \cline{2-9}
        & Euclid & 90 & 74 & 16 & 0 & 1 & 82.22\% & 100.00\% \\ \hline
        \multicolumn{9}{|c|}{UOBM-lite-10-36}   \\ \hline
       \multicolumn{2}{|c|}{Skep.}& 92 & 61 & 31 & 0 & 0 & 63.30\%  & 100\%\\ \hline
       \multicolumn{2}{|c|}{CMCS} & 92 & 71 & 12 & 5 & 4 & 77.17\% & 95.65\% \\ \hline
       \multicolumn{2}{|c|}{\#mc} & 92 & 71 & 12 & 5 & 4 & 77.17\% & 95.65\% \\ \hline 
       {Sentence-BERT} &  
       Cosine & 92 & 82 & 0 & 5 & 5 & 89.13\% & 94.57\%   \\ \cline{2-9}
        & Euclid & 92 & 86 & 0 & 5 & 1 & 93.48\% & 98.91\% \\
        \hline
       {ConSERT} & Cosine & 92 & 82 & 0 & 5 & 5 & 89.13\% & 94.57\% \\ \cline{2-9}
        & Euclid & 92 & 86 & 0 & 5 & 1 & 93.48\% & 98.91\% \\
        \hline
       {TransE} & Cosine   & 92 & 82 & 0 & 5 & 5 & 89.13\% & 94.57\% \\ \cline{2-9}
        & Euclid & 92 & 86 & 0 & 5 & 1 & 93.48\% & 98.91\%\\
        \hline
       {RDF2Vec} & Cosine & 92 & 86 & 0 & 5 & 1 & 93.48\% & 98.91\% \\ \cline{2-9}
        & Euclid & 92 & 71 & 12 & 5 & 4 & 77.17\% & 95.65\% \\ \hline
        \end{tabular} }
    \end{center}
     \caption{IA = Intended Answers, CA = Cautious Answers, RA = Reckless Answers, CIA = Counter-Intuitive Answers, IA Rate =Intended Answers(\%), ICR Rate = IA+CA+RA(\%)}
\end{table*}

\subsection{Experimental Settings}
The experiment includes four parts: generating the MCSs, transforming axioms with NaturalOWL or TripleOWL, scoring the MCSs and testing the queries. We perform the MCS generation system, NaturalOWL system and TripleOWL system in Java 1.8.0 environment, on a laptop with 1.80 GHz Intel core CPU, 16 GB RAM and a 64-bit Windows 10 operating system. The MCS generation system can be applied to any DLs. We implement the scoring function for selecting MCSs using Python 3.9.7 and PyTorch 1.11.0 with the same computer requirements as MCS generation system. For the query test experiment, we used the widely used ontology editor Protégé\footnote{https://protege.stanford.edu/} with version 5.5.0 and max memory set to 444MB. All the data and codes including MCS generator, NaturalOWL system, TripleOWL system, \textit{Injector} and scoring function are available at \href{https://anonymous.4open.science/r/Embedding-based-infer/}{https://anonymous.4open.science/r/Embedding-based-infer/}.

We selected three existing methods based on MCSs as baselines as follows.  \\
$\bullet$ Skeptical inference \cite{7}.\\
$\bullet$ CMCS: It is based on selecting those cardinality-maximal consistent subsets, known as a refinement on skeptical inference \cite{8}.\\
$\bullet$ \#mc: It is proposed by \cite{9}  considering the reliability of the information carried by the axioms as shown in Definitions 3, 4. 

Our proposed method is according to Definition 5, 6, 7 and we have implemented different variances corresponding to difference embedding models.

\cite{9} proposed a series of reasonable reasoning approaches based on MCS which satisfy a set of logical properties, but there is a lack of experiment to evaluate the performance of the reasoning methods. In this section, each algorithm is evaluated with respect to its effectiveness and efficiency. When reasoning with an inconsistent ontology, the effectiveness of one algorithm indicates the quality of the answers to the test queries. The efficiency of the algorithm is the time to generate answers to the test queries. We use the same evaluation method as \cite{iswc08, ijcai05}. We compare answers generated by the method to be evaluated with hand-crafted Gold Standard that contains the humanly-judged correct answer which is supposed by human,including ontology experts and postgraduate students. For a query, there might exist the following differences between an answer generated by the method to be evaluated and its intuitive answer. \\
$\bullet$ Intended Answer: the method’s answer is the same as the intuitive answer. \\
$\bullet$Counter-intuitive Answer: the method’s answer is opposite to the intuitive answer. Namely, the intuitive answer is `accepted’ whereas the method’s answer is `rejected’, or vice versa. \\
$\bullet$Cautious Answer: the intuitive answer is `accepted’ or `rejected’, but the method’s answer is `undetermined’. \\
$\bullet$Reckless Answer: the method’s answer is `accepted’ or `rejected’ whereas the intuitive answer is `undetermined’. Under this situation the method to be evaluated returns just one of the possible answers without seeking other possibly opposite answers, which may lead to `undetermined’.

We randomly select classes, individuals and properties to form the queries such as `Denmark(individual) Type country(concept) ?', `Administrator(concept) SubclassOf Person(concept) ?'. Especially, we give greater weight to classes, individuals and properties that are more likely to cause inconsistency when randomly forming queries. We compared these answers against a hand-crafted Gold Standard that contained the humanly-judged correct answer for all of these queries.

\vspace{-0.2cm}
\subsection{Evaluation Results}
Table 3 presents the evaluation results.  Baselines cannot give an effective MCS selection method because the semantic information of the ontology is ignored, which leads to poor reasoning quality. Compared with baselines, our method has a significant improvement in the intended answer (IA) rate. For AUT.cocus-edas, the IA rate of the baselines is lower than 7\%, but our proposed method based on sentence embedding can be above 96\%. The method based on knowledge graph embedding doesn't perform well here, because there is rich and important textual information in this ontology, which is not utilized in the knowledge-graph-embedding-based method, and simply concatenating parts of the triple can't represent the meaning of an axiom well especially for axioms with complex concepts.  For bioportal metadata and the two UOBM ontologies, our method has improved the IA rate by more than 10\% and 20\% respectively with comparison to the baselines.
The baselines achieve high ICR rate at the cost of low intended answer (IA) rate. However, our proposed method can achieve both high ICR rate and IA rate. For AUT.cocus-edas and bioportal metadata, most results for the ICR rate of our method are 100\%. For the two UOBM ontologies, our method can achieve more than 93\% and the majority of the results are above 98\% . These reflect the fact that the proposed method could provide promising results when reasoning with inconsistent ontologies. It owes to the semantic information of the axioms considered in our embedding-based method.



We also evaluate the efficiency of our method with respect to selecting MCSs. For the four ontologies in our experiment, the consumed time of the proposed method is within 10 minutes. Although the time consumed by the baselines is within several seconds and our proposed method spends more time than the baselines to score and select MCSs,  it is efficient enough in practice as the procedure of scoring and selection only needs to be performed once for each ontology and this process can be done offline. Suppose the selection is done, we also evaluate the time to execute each query. Our proposed approach is very efficient and a query can be answered within about half a second.

\vspace{-0.2cm}
\section{Conclusion and Discussion}
In this paper, we introduced a new approach to reasoning with inconsistent ontologies based on maximal consistent subsets. As far as we know, this is the first work that applies embedding techniques to  inconsistency-tolerant reasoning. In our work, we first proposed two methods for turning axioms into semantic vectors and computed the semantic similarity between axioms. We then proposed an approach for selecting the maximal consistent subsets and defined an inconsistency-tolerant inference relation. We showed the logical properties of proposed inconsistency-tolerant inference relation. We have proved that the inference relation we proposed satisfies the logic properties, which showed the rationality of the inference relation. We conducted extensive experiments on four ontologies and the experimental results show that our embedding-based method can outperform existing methods based on maximal consistent subsets. 

As for future work, we will give more careful consideration to dealing with complex axioms. We also plan to extend our method to inconsistency-tolerant inference with weighted ontologies and consider applying ChatGPT for translating axiom into sentences. Finally, we can explore applying embedding techniques to ontology repair by defining some relevant relations like \cite{35}.

\section{ACKNOWLEDGMENTS}
We thank the reviewers for their valuable comments.This work is partially supported by National Nature Science Foundation of China under No. U21A20488.

\bibliographystyle{ACM-Reference-Format}
\bibliography{sample-base.bib}


\begin{thebibliography}{45}


\ifx \showCODEN    \undefined \def \showCODEN     #1{\unskip}     \fi
\ifx \showDOI      \undefined \def \showDOI       #1{#1}\fi
\ifx \showISBNx    \undefined \def \showISBNx     #1{\unskip}     \fi
\ifx \showISBNxiii \undefined \def \showISBNxiii  #1{\unskip}     \fi
\ifx \showISSN     \undefined \def \showISSN      #1{\unskip}     \fi
\ifx \showLCCN     \undefined \def \showLCCN      #1{\unskip}     \fi
\ifx \shownote     \undefined \def \shownote      #1{#1}          \fi
\ifx \showarticletitle \undefined \def \showarticletitle #1{#1}   \fi
\ifx \showURL      \undefined \def \showURL       {\relax}        \fi
\providecommand\bibfield[2]{#2}
\providecommand\bibinfo[2]{#2}
\providecommand\natexlab[1]{#1}
\providecommand\showeprint[2][]{arXiv:#2}

\bibitem[Androutsopoulos et~al\mbox{.}(2013)]%
        {29}
\bibfield{author}{\bibinfo{person}{Ion Androutsopoulos},
  \bibinfo{person}{Gerasimos Lampouras}, {and} \bibinfo{person}{Dimitrios
  Galanis}.} \bibinfo{year}{2013}\natexlab{}.
\newblock \showarticletitle{Generating natural language descriptions from OWL
  ontologies: the NaturalOWL system}.
\newblock \bibinfo{journal}{\emph{Journal of Artificial Intelligence Research}}
   \bibinfo{volume}{48} (\bibinfo{year}{2013}), \bibinfo{pages}{671--715}.
\newblock


\bibitem[Baader et~al\mbox{.}(2003)]%
        {15}
\bibfield{author}{\bibinfo{person}{Franz Baader}, \bibinfo{person}{Diego
  Calvanese}, \bibinfo{person}{Deborah McGuinness}, \bibinfo{person}{Peter
  Patel-Schneider}, \bibinfo{person}{Daniele Nardi}, {et~al\mbox{.}}}
  \bibinfo{year}{2003}\natexlab{}.
\newblock \bibinfo{booktitle}{\emph{The description logic handbook: Theory,
  implementation and applications}}.
\newblock \bibinfo{publisher}{Cambridge university press}.
\newblock


\bibitem[Benferhat et~al\mbox{.}(1993)]%
        {8}
\bibfield{author}{\bibinfo{person}{Salem Benferhat}, \bibinfo{person}{Claudette
  Cayrol}, \bibinfo{person}{Didier Dubois}, \bibinfo{person}{J{\'e}r{\^o}me
  Lang}, {and} \bibinfo{person}{Henri Prade}.} \bibinfo{year}{1993}\natexlab{}.
\newblock \showarticletitle{Inconsistency management and prioritized
  syntax-based entailment}. In \bibinfo{booktitle}{\emph{IJCAI}},
  Vol.~\bibinfo{volume}{93}. \bibinfo{pages}{640--645}.
\newblock


\bibitem[Bienvenu(2020)]%
        {18}
\bibfield{author}{\bibinfo{person}{Meghyn Bienvenu}.}
  \bibinfo{year}{2020}\natexlab{}.
\newblock \showarticletitle{A short survey on inconsistency handling in
  ontology-mediated query answering}.
\newblock \bibinfo{journal}{\emph{KI-K{\"u}nstliche Intelligenz}}
  \bibinfo{volume}{34}, \bibinfo{number}{4} (\bibinfo{year}{2020}),
  \bibinfo{pages}{443--451}.
\newblock


\bibitem[Bienvenu and Bourgaux(2016)]%
        {17}
\bibfield{author}{\bibinfo{person}{Meghyn Bienvenu} {and}
  \bibinfo{person}{Camille Bourgaux}.} \bibinfo{year}{2016}\natexlab{}.
\newblock \showarticletitle{Inconsistency-tolerant querying of description
  logic knowledge bases}. In \bibinfo{booktitle}{\emph{Reasoning Web
  International Summer School}}. Springer, \bibinfo{pages}{156--202}.
\newblock


\bibitem[Bienvenu and Bourgaux(2022)]%
        {25}
\bibfield{author}{\bibinfo{person}{Meghyn Bienvenu} {and}
  \bibinfo{person}{Camille Bourgaux}.} \bibinfo{year}{2022}\natexlab{}.
\newblock \showarticletitle{Querying Inconsistent Prioritized Data with ORBITS:
  Algorithms, Implementation, and Experiments}.
\newblock \bibinfo{journal}{\emph{arXiv preprint arXiv:2202.07980}}
  (\bibinfo{year}{2022}).
\newblock


\bibitem[Bienvenu et~al\mbox{.}(2014)]%
        {22}
\bibfield{author}{\bibinfo{person}{Meghyn Bienvenu}, \bibinfo{person}{Camille
  Bourgaux}, {and} \bibinfo{person}{Fran{\c{c}}ois Goasdou{\'e}}.}
  \bibinfo{year}{2014}\natexlab{}.
\newblock \showarticletitle{Querying inconsistent description logic knowledge
  bases under preferred repair semantics}. In
  \bibinfo{booktitle}{\emph{Proceedings of the AAAI Conference on Artificial
  Intelligence}}, Vol.~\bibinfo{volume}{28}.
\newblock


\bibitem[Bienvenu et~al\mbox{.}(2019)]%
        {24}
\bibfield{author}{\bibinfo{person}{Meghyn Bienvenu}, \bibinfo{person}{Camille
  Bourgaux}, {and} \bibinfo{person}{Fran{\c{c}}ois Goasdou{\'e}}.}
  \bibinfo{year}{2019}\natexlab{}.
\newblock \showarticletitle{Computing and explaining query answers over
  inconsistent DL-Lite knowledge bases}.
\newblock \bibinfo{journal}{\emph{Journal of Artificial Intelligence Research}}
   \bibinfo{volume}{64} (\bibinfo{year}{2019}), \bibinfo{pages}{563--644}.
\newblock


\bibitem[Bordes et~al\mbox{.}(2013)]%
        {transe}
\bibfield{author}{\bibinfo{person}{Antoine Bordes}, \bibinfo{person}{Nicolas
  Usunier}, \bibinfo{person}{Alberto Garcia-Duran}, \bibinfo{person}{Jason
  Weston}, {and} \bibinfo{person}{Oksana Yakhnenko}.}
  \bibinfo{year}{2013}\natexlab{}.
\newblock \showarticletitle{Translating embeddings for modeling
  multi-relational data}.
\newblock \bibinfo{journal}{\emph{Advances in neural information processing
  systems}}  \bibinfo{volume}{26} (\bibinfo{year}{2013}).
\newblock


\bibitem[Chen et~al\mbox{.}(2020)]%
        {1}
\bibfield{author}{\bibinfo{person}{Xiaojun Chen}, \bibinfo{person}{Shengbin
  Jia}, {and} \bibinfo{person}{Yang Xiang}.} \bibinfo{year}{2020}\natexlab{}.
\newblock \showarticletitle{A review: Knowledge reasoning over knowledge
  graph}.
\newblock \bibinfo{journal}{\emph{Expert Systems with Applications}}
  \bibinfo{volume}{141} (\bibinfo{year}{2020}), \bibinfo{pages}{112948}.
\newblock


\bibitem[Devlin et~al\mbox{.}(2018)]%
        {11}
\bibfield{author}{\bibinfo{person}{Jacob Devlin}, \bibinfo{person}{Ming-Wei
  Chang}, \bibinfo{person}{Kenton Lee}, {and} \bibinfo{person}{Kristina
  Toutanova}.} \bibinfo{year}{2018}\natexlab{}.
\newblock \showarticletitle{BERT: Pre-training of deep bidirectional
  transformers for language understanding}.
\newblock \bibinfo{journal}{\emph{arXiv preprint arXiv:1810.04805}}
  (\bibinfo{year}{2018}).
\newblock


\bibitem[Du et~al\mbox{.}(2013)]%
        {21}
\bibfield{author}{\bibinfo{person}{Jianfeng Du}, \bibinfo{person}{Guilin Qi},
  {and} \bibinfo{person}{Yi-Dong Shen}.} \bibinfo{year}{2013}\natexlab{}.
\newblock \showarticletitle{Weight-based consistent query answering over
  inconsistent SHIQ knowledge bases}.
\newblock \bibinfo{journal}{\emph{Knowledge and Information Systems}}
  \bibinfo{volume}{34}, \bibinfo{number}{2} (\bibinfo{year}{2013}),
  \bibinfo{pages}{335--371}.
\newblock


\bibitem[Guo et~al\mbox{.}(2005)]%
        {31}
\bibfield{author}{\bibinfo{person}{Yuanbo Guo}, \bibinfo{person}{Zhengxiang
  Pan}, {and} \bibinfo{person}{Jeff Heflin}.} \bibinfo{year}{2005}\natexlab{}.
\newblock \showarticletitle{LUBM: A benchmark for OWL knowledge base systems}.
\newblock \bibinfo{journal}{\emph{Journal of Web Semantics}}
  \bibinfo{volume}{3}, \bibinfo{number}{2-3} (\bibinfo{year}{2005}),
  \bibinfo{pages}{158--182}.
\newblock


\bibitem[Haase et~al\mbox{.}(2005)]%
        {4}
\bibfield{author}{\bibinfo{person}{Peter Haase}, \bibinfo{person}{Frank~van
  Harmelen}, \bibinfo{person}{Zhisheng Huang}, \bibinfo{person}{Heiner
  Stuckenschmidt}, {and} \bibinfo{person}{York Sure}.}
  \bibinfo{year}{2005}\natexlab{}.
\newblock \showarticletitle{A framework for handling inconsistency in changing
  ontologies}. In \bibinfo{booktitle}{\emph{International semantic web
  conference}}. Springer, \bibinfo{pages}{353--367}.
\newblock


\bibitem[Hameed et~al\mbox{.}(2004)]%
        {3}
\bibfield{author}{\bibinfo{person}{Adil Hameed}, \bibinfo{person}{Alun Preece},
  {and} \bibinfo{person}{Derek Sleeman}.} \bibinfo{year}{2004}\natexlab{}.
\newblock \showarticletitle{Ontology reconciliation}.
\newblock In \bibinfo{booktitle}{\emph{Handbook on ontologies}}.
  \bibinfo{publisher}{Springer}, \bibinfo{pages}{231--250}.
\newblock


\bibitem[Han et~al\mbox{.}(2018)]%
        {openke}
\bibfield{author}{\bibinfo{person}{Xu Han}, \bibinfo{person}{Shulin Cao},
  \bibinfo{person}{Lv Xin}, \bibinfo{person}{Yankai Lin},
  \bibinfo{person}{Zhiyuan Liu}, \bibinfo{person}{Maosong Sun}, {and}
  \bibinfo{person}{Juanzi Li}.} \bibinfo{year}{2018}\natexlab{}.
\newblock \showarticletitle{OpenKE: An Open Toolkit for Knowledge Embedding}.
  In \bibinfo{booktitle}{\emph{Proceedings of EMNLP}}.
\newblock


\bibitem[Hao et~al\mbox{.}(2019)]%
        {joie}
\bibfield{author}{\bibinfo{person}{Junheng Hao}, \bibinfo{person}{Muhao Chen},
  \bibinfo{person}{Wenchao Yu}, \bibinfo{person}{Yizhou Sun}, {and}
  \bibinfo{person}{Wei Wang}.} \bibinfo{year}{2019}\natexlab{}.
\newblock \showarticletitle{Universal representation learning of knowledge
  bases by jointly embedding instances and ontological concepts}. In
  \bibinfo{booktitle}{\emph{Proceedings of the 25th ACM SIGKDD International
  Conference on Knowledge Discovery \& Data Mining}}.
  \bibinfo{pages}{1709--1719}.
\newblock


\bibitem[He et~al\mbox{.}(2022)]%
        {12}
\bibfield{author}{\bibinfo{person}{Yuan He}, \bibinfo{person}{Jiaoyan Chen},
  \bibinfo{person}{Denvar Antonyrajah}, {and} \bibinfo{person}{Ian Horrocks}.}
  \bibinfo{year}{2022}\natexlab{}.
\newblock \showarticletitle{BERTMap: a BERT-based ontology alignment system}.
  In \bibinfo{booktitle}{\emph{Proceedings of the AAAI Conference on Artificial
  Intelligence}}, Vol.~\bibinfo{volume}{36}. \bibinfo{pages}{5684--5691}.
\newblock


\bibitem[Hertling et~al\mbox{.}(2020)]%
        {13}
\bibfield{author}{\bibinfo{person}{Sven Hertling}, \bibinfo{person}{Jan
  Portisch}, {and} \bibinfo{person}{Heiko Paulheim}.}
  \bibinfo{year}{2020}\natexlab{}.
\newblock \showarticletitle{Supervised ontology and instance matching with
  MELT}.
\newblock \bibinfo{journal}{\emph{arXiv preprint arXiv:2009.11102}}
  (\bibinfo{year}{2020}).
\newblock


\bibitem[Huang and Harmelen(2008)]%
        {20}
\bibfield{author}{\bibinfo{person}{Zhisheng Huang} {and}
  \bibinfo{person}{Frank~van Harmelen}.} \bibinfo{year}{2008}\natexlab{}.
\newblock \showarticletitle{Using semantic distances for reasoning with
  inconsistent ontologies}. In \bibinfo{booktitle}{\emph{International Semantic
  Web Conference}}. Springer, \bibinfo{pages}{178--194}.
\newblock


\bibitem[Huang et~al\mbox{.}(2005a)]%
        {ijcai05}
\bibfield{author}{\bibinfo{person}{Zhisheng Huang}, \bibinfo{person}{F.~V.
  Harmelen}, {and} \bibinfo{person}{Annette ten Teije}.}
  \bibinfo{year}{2005}\natexlab{a}.
\newblock \showarticletitle{Reasoning with Inconsistent Ontologies}. In
  \bibinfo{booktitle}{\emph{International Joint Conference on Artificial
  Intelligence}}.
\newblock


\bibitem[Huang and van Harmelen(2008)]%
        {iswc08}
\bibfield{author}{\bibinfo{person}{Zhisheng Huang} {and} \bibinfo{person}{Frank
  van Harmelen}.} \bibinfo{year}{2008}\natexlab{}.
\newblock \showarticletitle{Using semantic distances for reasoning with
  inconsistent ontologies}. In \bibinfo{booktitle}{\emph{The Semantic Web-ISWC
  2008: 7th International Semantic Web Conference, ISWC 2008, Karlsruhe,
  Germany, October 26-30, 2008. Proceedings 7}}. Springer,
  \bibinfo{pages}{178--194}.
\newblock


\bibitem[Huang et~al\mbox{.}(2005b)]%
        {19}
\bibfield{author}{\bibinfo{person}{Zhisheng Huang}, \bibinfo{person}{Frank
  Van~Harmelen}, {and} \bibinfo{person}{Annette~Ten Teije}.}
  \bibinfo{year}{2005}\natexlab{b}.
\newblock \showarticletitle{Reasoning with inconsistent ontologies}. In
  \bibinfo{booktitle}{\emph{Proceedings of the 19th International Joint
  Conference on Artificial Intelligence}}. \bibinfo{pages}{454--459}.
\newblock


\bibitem[Ji et~al\mbox{.}(2014)]%
        {kbs14}
\bibfield{author}{\bibinfo{person}{Qiu Ji}, \bibinfo{person}{Zhiqiang Gao},
  \bibinfo{person}{Zhisheng Huang}, {and} \bibinfo{person}{Man Zhu}.}
  \bibinfo{year}{2014}\natexlab{}.
\newblock \showarticletitle{Measuring effectiveness of ontology debugging
  systems}.
\newblock \bibinfo{journal}{\emph{Knowledge-Based Systems}}
  \bibinfo{volume}{71} (\bibinfo{year}{2014}), \bibinfo{pages}{169--186}.
\newblock


\bibitem[Ji et~al\mbox{.}(2009)]%
        {35}
\bibfield{author}{\bibinfo{person}{Qiu Ji}, \bibinfo{person}{Guilin Qi}, {and}
  \bibinfo{person}{Peter Haase}.} \bibinfo{year}{2009}\natexlab{}.
\newblock \showarticletitle{A relevance-directed algorithm for finding
  justifications of DL entailments}. In \bibinfo{booktitle}{\emph{Asian
  Semantic Web Conference}}. Springer, \bibinfo{pages}{306--320}.
\newblock


\bibitem[Konieczny et~al\mbox{.}(2019)]%
        {9}
\bibfield{author}{\bibinfo{person}{Sébastien Konieczny},
  \bibinfo{person}{Pierre Marquis}, {and} \bibinfo{person}{Srdjan Vesic}.}
  \bibinfo{year}{2019}\natexlab{}.
\newblock \showarticletitle{Rational Inference Relations from Maximal
  Consistent Subsets Selection}. In \bibinfo{booktitle}{\emph{Proceedings of
  the Twenty-Eighth International Joint Conference on Artificial Intelligence,
  {IJCAI-19}}}. \bibinfo{publisher}{International Joint Conferences on
  Artificial Intelligence Organization}, \bibinfo{pages}{1749--1755}.
\newblock
\urldef\tempurl%
\url{https://doi.org/10.24963/ijcai.2019/242}
\showDOI{\tempurl}


\bibitem[Kraus et~al\mbox{.}(1990)]%
        {7}
\bibfield{author}{\bibinfo{person}{Sarit Kraus}, \bibinfo{person}{Daniel
  Lehmann}, {and} \bibinfo{person}{Menachem Magidor}.}
  \bibinfo{year}{1990}\natexlab{}.
\newblock \showarticletitle{Nonmonotonic reasoning, preferential models and
  cumulative logics}.
\newblock \bibinfo{journal}{\emph{Artificial intelligence}}
  \bibinfo{volume}{44}, \bibinfo{number}{1-2} (\bibinfo{year}{1990}),
  \bibinfo{pages}{167--207}.
\newblock


\bibitem[Liu et~al\mbox{.}(2019)]%
        {2}
\bibfield{author}{\bibinfo{person}{Jin Liu}, \bibinfo{person}{Xin Zhang},
  \bibinfo{person}{Yunhui Li}, \bibinfo{person}{Jin Wang}, {and}
  \bibinfo{person}{Hye-Jin Kim}.} \bibinfo{year}{2019}\natexlab{}.
\newblock \showarticletitle{Deep learning-based reasoning with multi-ontology
  for IoT applications}.
\newblock \bibinfo{journal}{\emph{IEEE Access}}  \bibinfo{volume}{7}
  (\bibinfo{year}{2019}), \bibinfo{pages}{124688--124701}.
\newblock


\bibitem[Ma et~al\mbox{.}(2006)]%
        {32}
\bibfield{author}{\bibinfo{person}{Li Ma}, \bibinfo{person}{Yang Yang},
  \bibinfo{person}{Zhaoming Qiu}, \bibinfo{person}{Guotong Xie},
  \bibinfo{person}{Yue Pan}, {and} \bibinfo{person}{Shengping Liu}.}
  \bibinfo{year}{2006}\natexlab{}.
\newblock \showarticletitle{Towards a complete OWL ontology benchmark}. In
  \bibinfo{booktitle}{\emph{European Semantic Web Conference}}. Springer,
  \bibinfo{pages}{125--139}.
\newblock


\bibitem[Ma et~al\mbox{.}(2007)]%
        {26}
\bibfield{author}{\bibinfo{person}{Yue Ma}, \bibinfo{person}{Pascal Hitzler},
  {and} \bibinfo{person}{Zuoquan Lin}.} \bibinfo{year}{2007}\natexlab{}.
\newblock \showarticletitle{Algorithms for paraconsistent reasoning with OWL}.
  In \bibinfo{booktitle}{\emph{European Semantic Web Conference}}. Springer,
  \bibinfo{pages}{399--413}.
\newblock


\bibitem[Maier et~al\mbox{.}(2013)]%
        {27}
\bibfield{author}{\bibinfo{person}{Frederick Maier}, \bibinfo{person}{Yue Ma},
  {and} \bibinfo{person}{Pascal Hitzler}.} \bibinfo{year}{2013}\natexlab{}.
\newblock \showarticletitle{Paraconsistent OWL and related logics}.
\newblock \bibinfo{journal}{\emph{Semantic Web}} \bibinfo{volume}{4},
  \bibinfo{number}{4} (\bibinfo{year}{2013}), \bibinfo{pages}{395--427}.
\newblock


\bibitem[Mikolov et~al\mbox{.}(2013)]%
        {word2vec}
\bibfield{author}{\bibinfo{person}{Tomas Mikolov}, \bibinfo{person}{Kai Chen},
  \bibinfo{person}{Greg Corrado}, {and} \bibinfo{person}{Jeffrey Dean}.}
  \bibinfo{year}{2013}\natexlab{}.
\newblock \bibinfo{title}{Efficient Estimation of Word Representations in
  Vector Space}.
\newblock
\newblock
\showeprint{arXiv:1301.3781}


\bibitem[Mishra and Viradiya(2019)]%
        {sesurvey}
\bibfield{author}{\bibinfo{person}{Mridul~K Mishra} {and}
  \bibinfo{person}{Jaydeep Viradiya}.} \bibinfo{year}{2019}\natexlab{}.
\newblock \showarticletitle{Survey of sentence embedding methods}.
\newblock \bibinfo{journal}{\emph{International Journal of Applied Science and
  Computations}} \bibinfo{volume}{6}, \bibinfo{number}{3}
  (\bibinfo{year}{2019}), \bibinfo{pages}{592--592}.
\newblock


\bibitem[Nickel et~al\mbox{.}(2016)]%
        {hole}
\bibfield{author}{\bibinfo{person}{Maximilian Nickel}, \bibinfo{person}{Lorenzo
  Rosasco}, {and} \bibinfo{person}{Tomaso Poggio}.}
  \bibinfo{year}{2016}\natexlab{}.
\newblock \showarticletitle{Holographic embeddings of knowledge graphs}. In
  \bibinfo{booktitle}{\emph{Proceedings of the AAAI Conference on Artificial
  Intelligence}}, Vol.~\bibinfo{volume}{30}.
\newblock


\bibitem[Reimers and Gurevych(2019)]%
        {30}
\bibfield{author}{\bibinfo{person}{Nils Reimers} {and} \bibinfo{person}{Iryna
  Gurevych}.} \bibinfo{year}{2019}\natexlab{}.
\newblock \showarticletitle{Sentence-bert: Sentence embeddings using siamese
  bert-networks}.
\newblock \bibinfo{journal}{\emph{arXiv preprint arXiv:1908.10084}}
  (\bibinfo{year}{2019}).
\newblock


\bibitem[Rescher and Manor(1970)]%
        {6}
\bibfield{author}{\bibinfo{person}{Nicholas Rescher} {and}
  \bibinfo{person}{Ruth Manor}.} \bibinfo{year}{1970}\natexlab{}.
\newblock \showarticletitle{On inference from inconsistent premisses}.
\newblock \bibinfo{journal}{\emph{Theory and decision}} \bibinfo{volume}{1},
  \bibinfo{number}{2} (\bibinfo{year}{1970}), \bibinfo{pages}{179--217}.
\newblock


\bibitem[Ristoski et~al\mbox{.}(2017)]%
        {14}
\bibfield{author}{\bibinfo{person}{Petar Ristoski}, \bibinfo{person}{Stefano
  Faralli}, \bibinfo{person}{Simone~Paolo Ponzetto}, {and}
  \bibinfo{person}{Heiko Paulheim}.} \bibinfo{year}{2017}\natexlab{}.
\newblock \showarticletitle{Large-scale taxonomy induction using entity and
  word embeddings}. In \bibinfo{booktitle}{\emph{Proceedings of the
  International Conference on Web Intelligence}}. \bibinfo{pages}{81--87}.
\newblock


\bibitem[Ristoski and Paulheim(2016)]%
        {rdf2vec}
\bibfield{author}{\bibinfo{person}{Petar Ristoski} {and} \bibinfo{person}{Heiko
  Paulheim}.} \bibinfo{year}{2016}\natexlab{}.
\newblock \showarticletitle{Rdf2vec: Rdf graph embeddings for data mining}. In
  \bibinfo{booktitle}{\emph{International Semantic Web Conference}}. Springer,
  \bibinfo{pages}{498--514}.
\newblock


\bibitem[Schlobach et~al\mbox{.}(2003)]%
        {5}
\bibfield{author}{\bibinfo{person}{Stefan Schlobach}, \bibinfo{person}{Ronald
  Cornet}, {et~al\mbox{.}}} \bibinfo{year}{2003}\natexlab{}.
\newblock \showarticletitle{Non-standard reasoning services for the debugging
  of description logic terminologies}. In \bibinfo{booktitle}{\emph{Ijcai}},
  Vol.~\bibinfo{volume}{3}. \bibinfo{pages}{355--362}.
\newblock


\bibitem[Tsalapati et~al\mbox{.}(2016)]%
        {23}
\bibfield{author}{\bibinfo{person}{Eleni Tsalapati}, \bibinfo{person}{Giorgos
  Stoilos}, \bibinfo{person}{Giorgos~B Stamou}, {and} \bibinfo{person}{George
  Koletsos}.} \bibinfo{year}{2016}\natexlab{}.
\newblock \showarticletitle{Efficient Query Answering over Expressive
  Inconsistent Description Logics.}. In \bibinfo{booktitle}{\emph{IJCAI}}.
  \bibinfo{pages}{1279--1285}.
\newblock


\bibitem[Vandewiele et~al\mbox{.}(2022)]%
        {pyrdf2vec}
\bibfield{author}{\bibinfo{person}{Gilles Vandewiele}, \bibinfo{person}{Bram
  Steenwinckel}, \bibinfo{person}{Terencio Agozzino}, {and}
  \bibinfo{person}{Femke Ongenae}.} \bibinfo{year}{2022}\natexlab{}.
\newblock \showarticletitle{pyRDF2Vec: A Python Implementation and Extension of
  RDF2Vec}.
\newblock  (\bibinfo{year}{2022}).
\newblock
\urldef\tempurl%
\url{https://doi.org/10.48550/ARXIV.2205.02283}
\showDOI{\tempurl}


\bibitem[Wang et~al\mbox{.}(2017)]%
        {kge}
\bibfield{author}{\bibinfo{person}{Quan Wang}, \bibinfo{person}{Zhendong Mao},
  \bibinfo{person}{Bin Wang}, {and} \bibinfo{person}{Li Guo}.}
  \bibinfo{year}{2017}\natexlab{}.
\newblock \showarticletitle{Knowledge Graph Embedding: A Survey of Approaches
  and Applications}.
\newblock \bibinfo{journal}{\emph{IEEE Transactions on Knowledge and Data
  Engineering}} \bibinfo{volume}{29}, \bibinfo{number}{12}
  (\bibinfo{year}{2017}), \bibinfo{pages}{2724--2743}.
\newblock
\urldef\tempurl%
\url{https://doi.org/10.1109/TKDE.2017.2754499}
\showDOI{\tempurl}


\bibitem[Yan et~al\mbox{.}(2021)]%
        {consert}
\bibfield{author}{\bibinfo{person}{Yuanmeng Yan}, \bibinfo{person}{Rumei Li},
  \bibinfo{person}{Sirui Wang}, \bibinfo{person}{Fuzheng Zhang},
  \bibinfo{person}{Wei Wu}, {and} \bibinfo{person}{Weiran Xu}.}
  \bibinfo{year}{2021}\natexlab{}.
\newblock \showarticletitle{Consert: A contrastive framework for
  self-supervised sentence representation transfer}.
\newblock \bibinfo{journal}{\emph{arXiv preprint arXiv:2105.11741}}
  (\bibinfo{year}{2021}).
\newblock


\bibitem[Yang et~al\mbox{.}(2014)]%
        {distmult}
\bibfield{author}{\bibinfo{person}{Bishan Yang}, \bibinfo{person}{Wen-tau Yih},
  \bibinfo{person}{Xiaodong He}, \bibinfo{person}{Jianfeng Gao}, {and}
  \bibinfo{person}{Li Deng}.} \bibinfo{year}{2014}\natexlab{}.
\newblock \showarticletitle{Embedding entities and relations for learning and
  inference in knowledge bases}.
\newblock \bibinfo{journal}{\emph{arXiv preprint arXiv:1412.6575}}
  (\bibinfo{year}{2014}).
\newblock


\bibitem[Zhang et~al\mbox{.}(2014)]%
        {28}
\bibfield{author}{\bibinfo{person}{Xiaowang Zhang}, \bibinfo{person}{Guohui
  Xiao}, \bibinfo{person}{Zuoquan Lin}, {and} \bibinfo{person}{Jan Van~den
  Bussche}.} \bibinfo{year}{2014}\natexlab{}.
\newblock \showarticletitle{Inconsistency-tolerant reasoning with OWL DL}.
\newblock \bibinfo{journal}{\emph{International Journal of Approximate
  Reasoning}} \bibinfo{volume}{55}, \bibinfo{number}{2} (\bibinfo{year}{2014}),
  \bibinfo{pages}{557--584}.
\newblock


\end{thebibliography}

\end{document}